\documentclass[10pt,twocolumn,letterpaper]{article}

\usepackage{cvpr}              

\usepackage{colortbl}

\definecolor{cvprblue}{rgb}{0.21,0.49,0.74}
\usepackage[pagebackref,breaklinks,colorlinks,allcolors=cvprblue]{hyperref}

\def\paperID{*****} 
\def\confName{CVPR}
\def\confYear{2025}

\title{CTRL-GS: Cascaded Temporal Residue Learning for 4D Gaussian Splatting}

\author{Karly Hou, Wanhua Li, Hanspeter Pfister\\
Harvard University\\
{\tt\small karly@alumni.harvard.edu, wanhuali@g.harvard.edu, pfister@seas.harvard.edu}
}

\begin{document}
\maketitle
\begin{figure*}[htb]
\centering
\includegraphics[width=\textwidth]{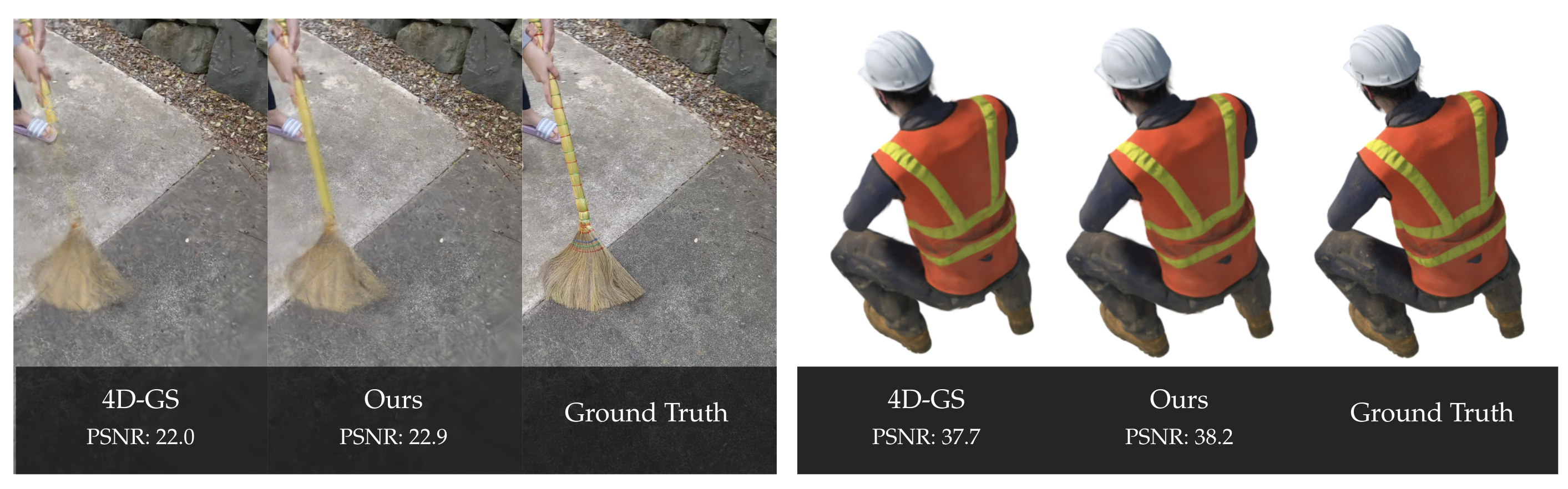}  
\caption{Our method achieves improved reconstruction accuracy for dynamic scenes at high image resolutions across settings. Current models tend to perform poorly on scenes with high motion, occluded areas, and fine details. \textbf{Left:} CTRL-GS correctly constructs the broom handle and more accurate shading details, where the 4D-GS reconstruction is blurry and omits the handle. \textbf{Right:} CTRL-GS reproduces fine details with accurate structure and depth, where the 4D-GS reconstruction omits fine textures and lighting changes.}
\label{fig:title}
\end{figure*}

\begin{abstract}
Recently, Gaussian Splatting methods have emerged as a desirable substitute for prior Radiance Field methods for novel-view synthesis of scenes captured with multi-view images or videos. In this work, we propose a novel extension to 4D Gaussian Splatting for dynamic scenes. Drawing on ideas from residual learning, we hierarchically decompose the dynamic scene into a "video-segment-frame" structure, with segments dynamically adjusted by optical flow. Then, instead of directly predicting the time-dependent signals, we model the signal as the sum of video-constant values, segment-constant values, and frame-specific residuals, as inspired by the success of residual learning. This approach allows more flexible models that adapt to highly variable scenes. We demonstrate state-of-the-art visual quality and real-time rendering on several established datasets, with the greatest improvements on complex scenes with large movements, occlusions, and fine details, where current methods degrade most.
\end{abstract}    
\section{Introduction}
\label{sec:intro}

Novel view synthesis is an important task for applications including medical imaging, autonomous driving, robotics, AR/VR, animation and gaming, and more. However, representing and rendering dynamic scenes with high visual accuracy and efficiency is challenging, especially for scenes with complex geometry and motion.

In recent years, Neural Radiance Fields (NeRFs) \cite{MST+20} and their extensions have achieved high-quality novel view synthesis results by using implicit radiance fields to represent light distribution without explicit scene geometry definition. Point radiances are computed on-the-fly by prompting the neural network rather than explicitly stored, allowing differentiable and compact representations of complex scenes, but training and rendering can be slow due to volumetric ray marching.

Kerbl and Kopanas et al. \cite{KKL+23} introduced 3D Gaussian Splatting (3D-GS) as an alternative to prior radiance fields, aiming to achieve high visual quality for unbounded and complete static scenes (as opposed to isolated objects) and render high-resolution images in real time. Explicit scene representation with 3D Gaussians allows flexible manipulation and differentiable splatting significantly boosts rendering speed from volumetric rendering. Splatting \cite{W89, LH91, LG95, MY96} involves projecting geometric primitives like points or discs onto an image plane to simulate the appearance of 3D objects. 

4D Gaussian Splatting (4D-GS) \cite{WYF+23} extends this work to dynamic scenes with moving objects by leveraging a spatial-temporal structure encoder, Gaussian deformation field, and deformation decoder. Instead of constructing 3D Gaussians at each timestamp, which is costly especially at scale, 4D-GS performs transformations on a canonical set of 3D Gaussians. 4D-GS achieves high-quality reconstructions while maintaining high rendering speed, but it often produces errors on scenes with large movements, occlusions, and fine details.

In this paper, we propose a novel extension to 4D Gaussian Splatting to address these challenges. We evaluate CTRL-GS on synthetic and real-world scenes with a range of types of motion. CTRL-GS achieves higher quantitative and qualitative reconstruction accuracy, with the largest improvements for challenging scenes with large movements. In summary, our main contributions include:

\begin{enumerate}
    \setlength{\itemsep}{0pt}  
    \item Three approaches to constructing temporal windows for spatial-temporal features.
    \item Temporally-local dynamic MLPs that model deformation fields over a subsegment of time.
    \item A hierarchical time signal modeling scheme that decomposes the dynamic scene into a "video-segment-frame residual" structure, offering both built-in scene continuity and flexibility for complex scenes.
\end{enumerate}

\section{Related Work}

\begin{figure*}
\centering
\includegraphics[width=\textwidth]{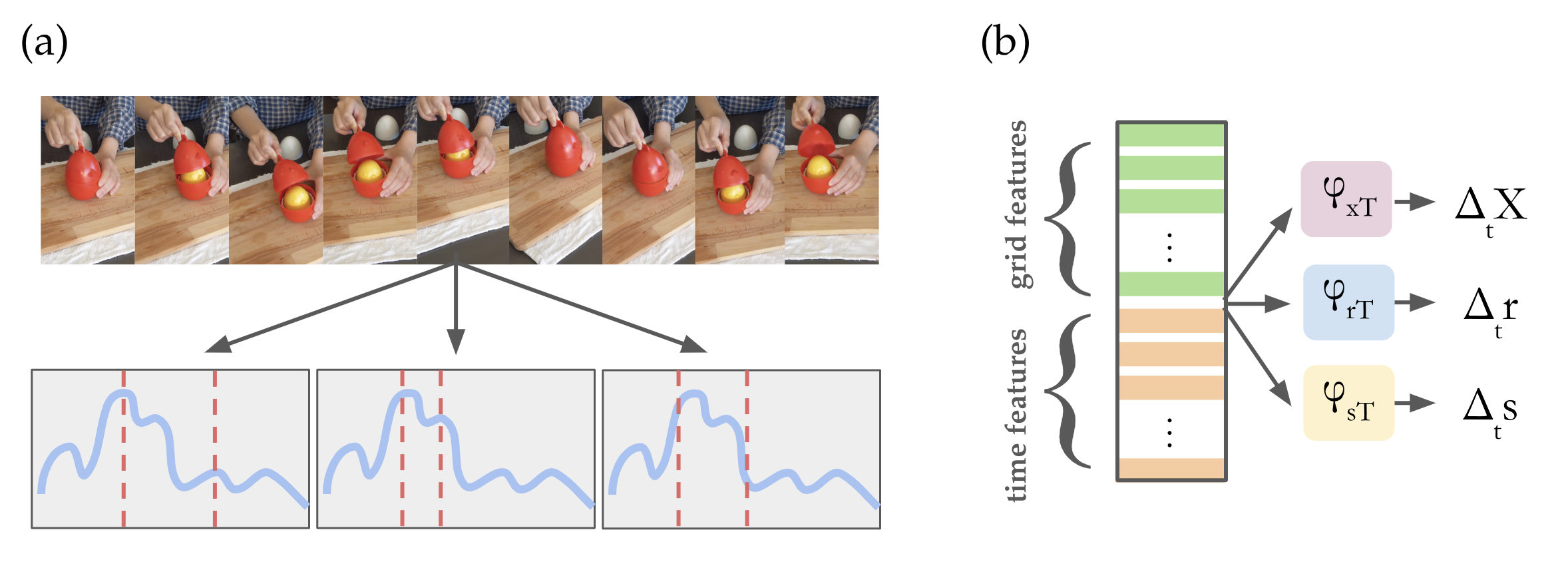}  
\caption{Two key components of our method. a) We divide the frames of the dynamic scene into temporal windows. We assess three different methods: 1) equal divisions, 2) divisions at the $N$ frame-pairs of highest optical flow, 3) divisions based on greedy optical flow thresholds. b) We construct MLPs $\varphi_{xT}, \varphi_{rT}, \varphi_{sT}$, which take encoded spatial-temporal features and produce intermediate mean components for position, rotation, and scaling for each temporal window individually. These components are combined with the original 3D Gaussians and point-in-time deformations to construct our final features.}
\label{fig:method}
\end{figure*}

\subsection{NeRF Methods}
Nerfies \cite{PSB+21} extends NeRF to optimize a deformation field per observation to handle non-rigidly deforming objects. HyperNeRF \cite{PSH+21} models more complex scene dynamics through a higher-dimension canonical space and a learned warp field to handle both geometric and temporal variations. FFDNeRF \cite{GSD+23} uses differentiable forward flow motion modeling for dynamic view synthesis. TiNeuVox-B \cite{FYW+22} represents scenes with time-aware voxel features and uses a multi-distance interpolation method to model both small and large motions. V4D \cite{GXH+23} uses 3D voxels to directly model 4D neural radiance fields without the need for a canonical space. K-Planes \cite{FMR+23} uses $d$ choose 2 planes to represent a $d$-dimension scene, allowing easy addition of dimension-specific priors. A variety of other works \cite{LSS+19, ZRS+20, BMT+21, LNS+21, BMV+22, FYT+22, CJ23, LGM+23} explore related ideas.

\subsection{Gaussian Splatting}
A variety of work builds on 3D-GS \cite{KKL+23} to achieve higher accuracy, faster rendering, more efficient representation, flexible manipulation, and diverse applications. \cite{FLK+24} demonstrates COLMAP-free 3D-GS. GauFRe \cite{LKL+25} constructs deformable models and separate static and dynamic components. DynMF \cite{KLD24} decomposes per-point motions in a dynamic scene into a small set of explicit trajectories. SC-GS \cite{HSY+24} uses sparse control points and dense Gaussians to enable user-controlled motion editing. PhysGaussian \cite{XZQ+24} integrates physical dynamics, combining 3D-GS with kinematic deformations and mechanical stress attributes. SG-Splatting \cite{WCY24} accelerates rendering speed with Spherical Gaussians. SwinGS \cite{SNS+23} uses temporally-local windows for dynamic 3D Gaussians. DreamGaussian4D \cite{RPT+23} refines generated 4D Gaussians with a pre-trained image-to-video diffusion model. GAGS \cite{PWL+24} incorporates CLIP features for scene understanding. 

Many Gaussian Splatting applications \cite{LTY+23, LZT+23, AYS+24, PSL+24, QWM+24,qin2024langsplat,li20254d} have been demonstrated, such as   \cite{SSS+24} for human and face models, \cite{ZSX+24} for urban and autonomous driving scenes, \cite{KKM+24} for SLAM, \cite{YZG+24} for reflective surfaces, and more.

\subsection{Residual Learning}
The original deep residual learning for image recognition paper \cite{HZR+16} introduces the idea that instead of approximating a target function $H(x)$ directly, a network can be designed to instead approximate a residual function $F(x)$ such that $H(x) = F(x)+x$. Though both forms may be able to asymptotically approximate the desired functions, the reformulation may be easier to learn. If additional layers can be constructed as identity mappings, then a deeper model should be able to perform at least as well as a shallower one; with the residual learning reformulation, the model can drive weights of nonlinear layers toward zero to approach identity mappings. If the optimal function is closer to an identity mapping than to zero mapping, it should be easier for the model to find perturbations with reference to an identity mapping than to learn the entire function. This idea inspires our hierarchical decomposition in a temporal setting.

\section{Method}

\begin{figure*}
\centering
\includegraphics[width=\textwidth]{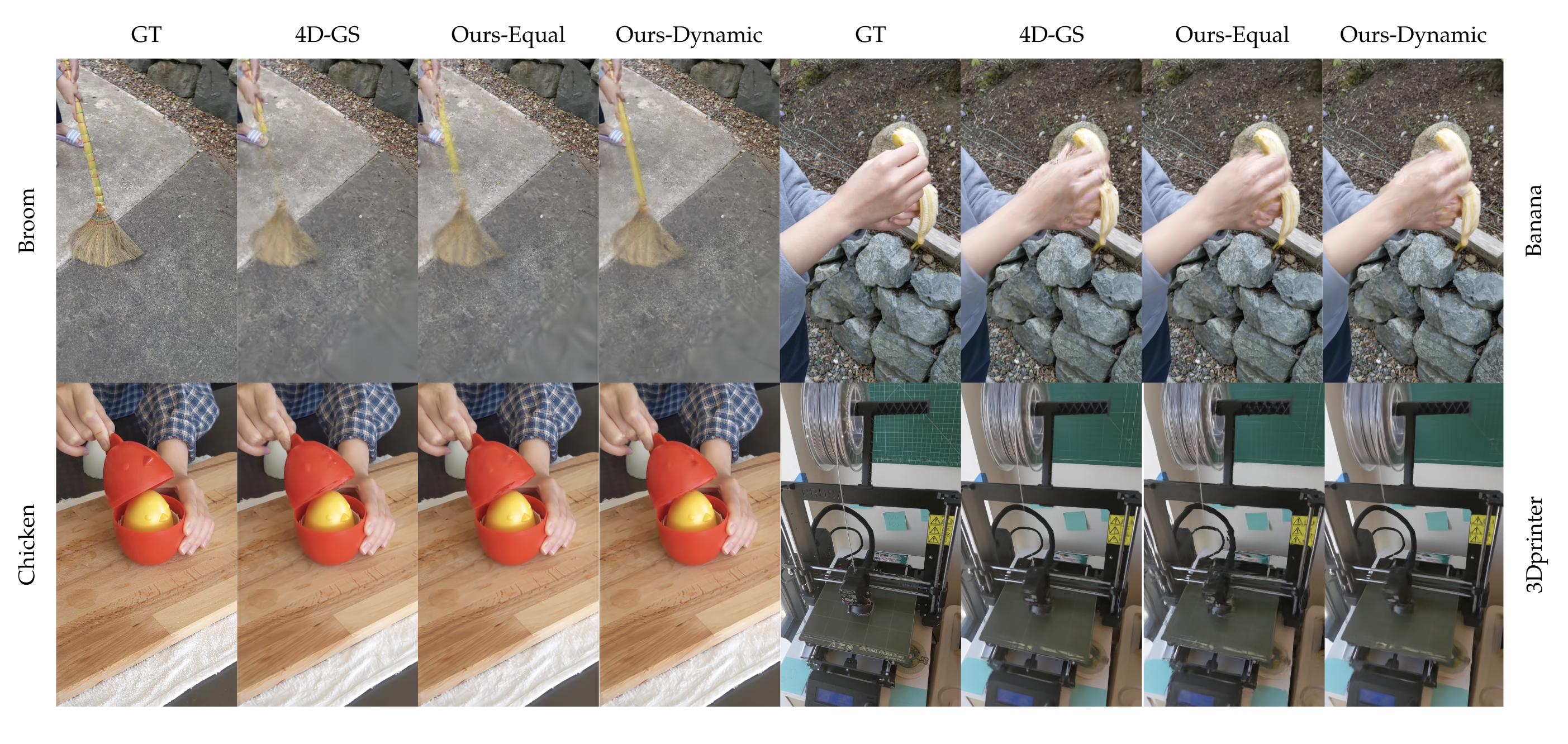}  
\caption{Comparisons between ground truth (GT), baseline 4D-GS \cite{WYF+23} result, and our method with equal and dynamic threshold-based windows on HyperNeRF \cite{PSH+21} validation rig scenes. \textbf{Broom:} CTRL-GS correctly constructs the broom handle and offers better depth shading. \textbf{Banana:} CTRL-GS correctly places fingers. \textbf{Chicken:} CTRL-GS preserves edge structural details and beak shading, and reduces blur. \textbf{3D Printer}: CTRL-GS does not produce improved results on this example.}
\label{fig:hypernerf_images}
\end{figure*}

\begin{figure*}
\centering
\includegraphics[width=\textwidth]{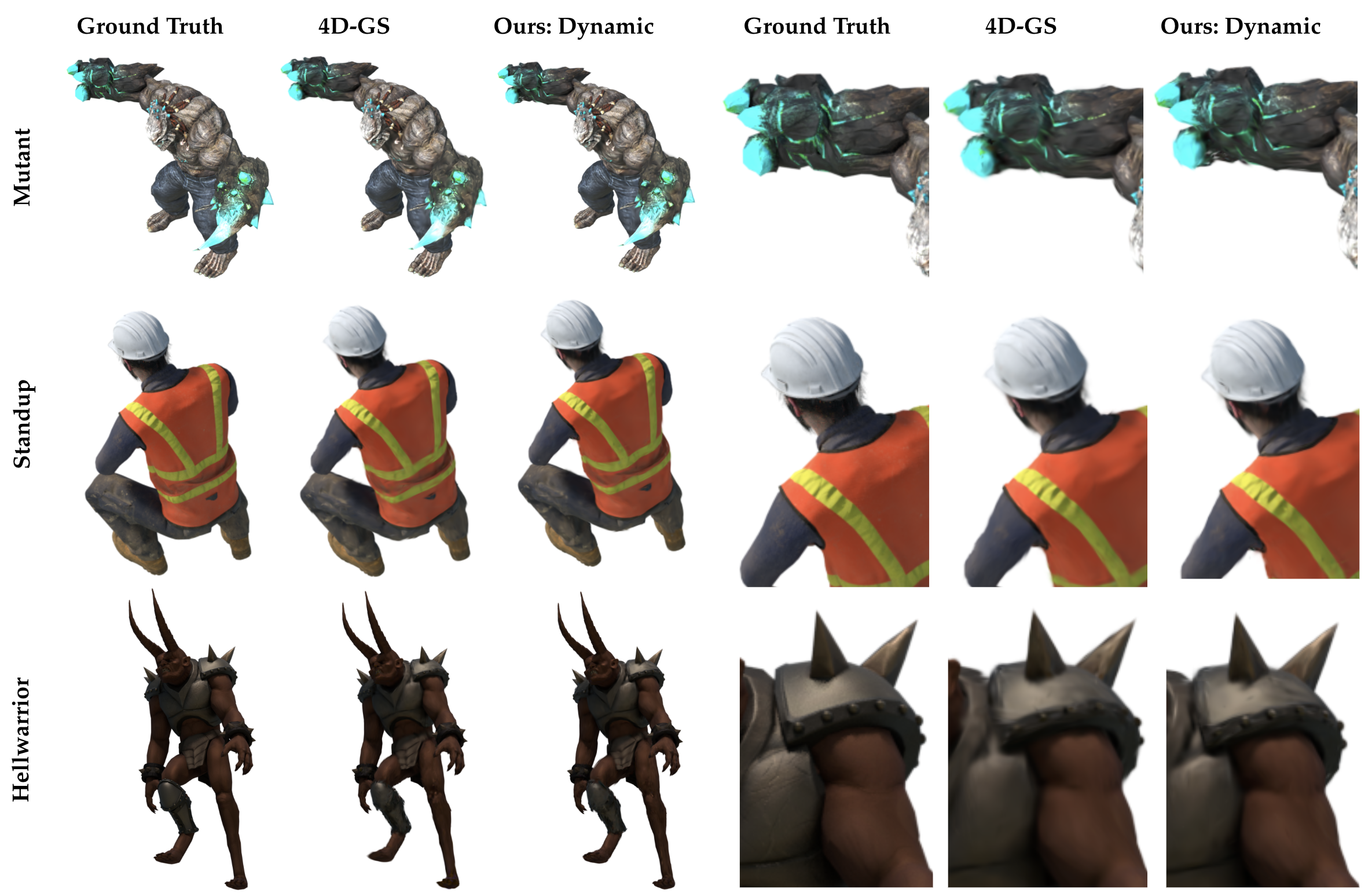}  
\caption{Comparisons between ground truth (GT), 4D-GS, and our method with dynamic threshold-based windows on D-NeRF \cite{PSH+21} synthetic scenes. CTRL-GS shows greater reconstruction fidelity and less blur, especially on fine details and sharp edges. \textbf{Mutant:} See blue claw structure, clarity of small blue veins, and placement of cracks. \textbf{Standup:} See helmet edges, helmet ridges, and vest wrinkles. \textbf{Hellwarrior:} See shoulder spheres, chestplate ridges, and luminosity of collar highlight.}
\label{fig:dnerf_images}
\end{figure*}

\begin{figure*}
\centering
\includegraphics[width=0.8\textwidth]{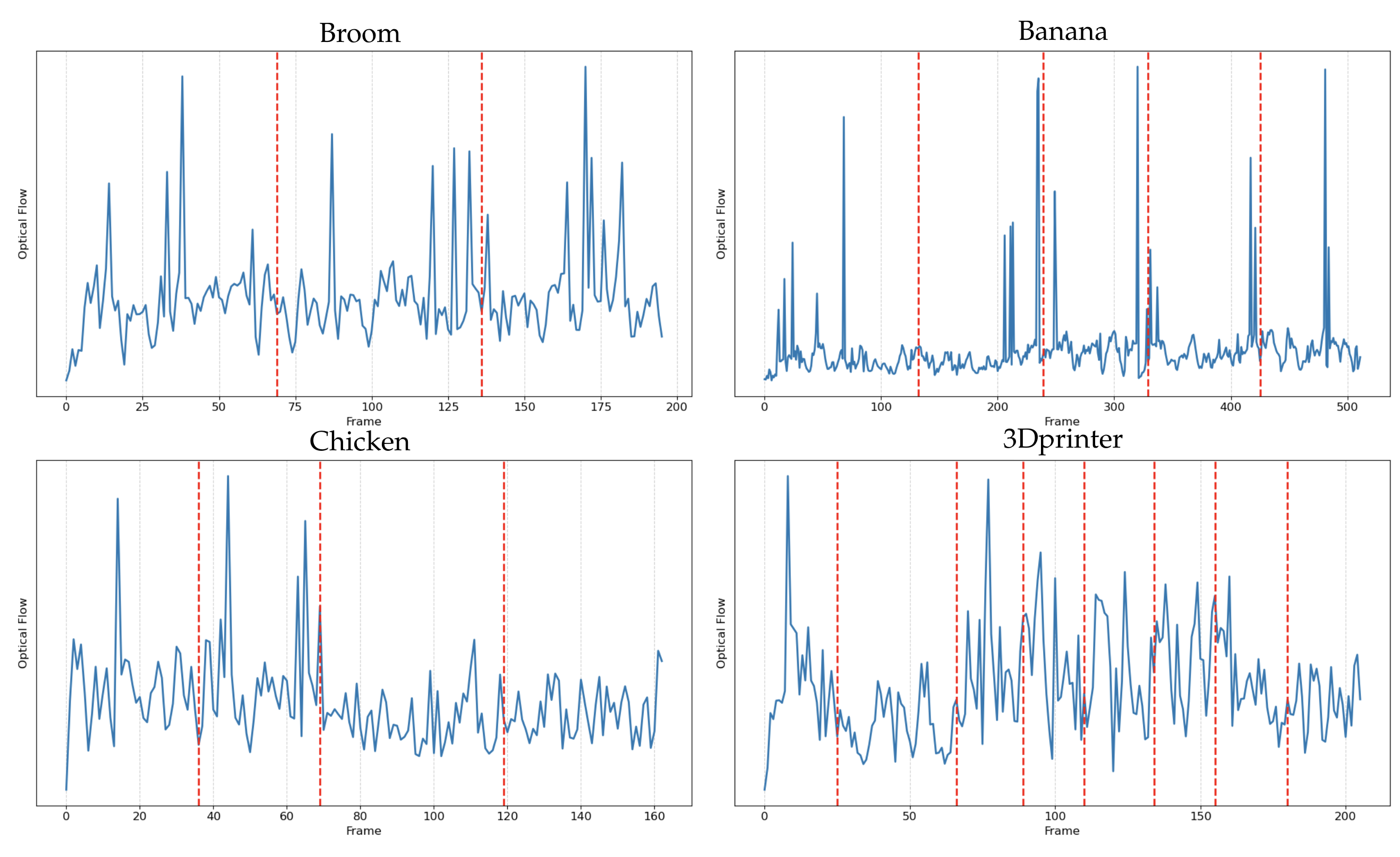}  
\caption{Magnitude of optical flow $f_{ij}$ over each of the HyperNeRF \cite{PSH+21} datasets a) \textsc{Broom}, b) \textsc{Banana}, c) \textsc{Chicken}, \textsc{3Dprinter}. Red vertical lines denote optimal temporal window segmentation when constructed with the dynamic thresholds method.}
\label{fig:time_graph}
\end{figure*}

\subsection{Preliminaries: 3D Gaussian Splatting}
Starting from a sparse set of SfM \cite{SF16} points, each 3D Gaussian is defined by covariance matrix $\Sigma$ in world space \cite{ZPV+01} centered at point $\mathcal{X}$: 
\begin{equation}
G(\mathcal{X}) = e^{-\frac{1}{2}\mathcal{X}^T\Sigma^{-1}\mathcal{X}}
\end{equation}

For rendering, $\Sigma$ is converted to a $\Sigma'$ covariance matrix represented in camera coordinates: 
\begin{equation}
\Sigma' = JW\Sigma W^TJ^T
\end{equation}
where $W$ is a viewing transformation matrix and $J$ is the Jacobian of the affine approximation of the projective transformation.

This covariance $\Sigma$ can be decomposed into scaling matrix $S$ and rotation matrix $R$ for optimization: 
\begin{equation}
\Sigma = RSS^TR^T
\end{equation}

Next, optimization of positions $\mathcal{X} \in \mathbb{R}^3$, color as defined by spherical harmonic (SH) coefficients $\mathcal{C} \in \mathbb{R}^k$ (where $k$ represents the number of SH functions), opacities $\alpha \in \mathbb{R}$, rotations $r \in \mathbb{R}^4$, and scaling factors $s \in \mathbb{R}^3$ is interleaved with steps that adaptively control density of the Gaussians.

Finally, a tile-based rasterizer filters out Gaussians unlikely to be visible in the viewing area, new instances of the Gaussians are sorted, and the $N$ points overlapping a pixel are blended using 
\begin{equation}
C = \sum_{i \in N}c_i\alpha_i \prod_{j=1}^{i-1}(1-\alpha_i)
\end{equation}
where $c_i$ and $\alpha_i$ represent the color and opacity of each point.

\subsection{Preliminaries: 4D Gaussian Splatting}
4D-GS \cite{WYF+23} leverages a Gaussian deformation field network $\mathcal{F}$, spatial-temporal structure encoder $\mathcal{H}$, and multi-head deformation decoder $\mathcal{D}$ to produce deformed 3D Gaussians $\mathcal{G}'$ from the original 3D Gaussians $\mathcal{G}$.

First, $\mathcal{H}$ including a multi-resolution HexPlane $R(i,j)$ and a tiny MLP $\phi_d$ encode spatial-temporal features: $\mathcal{H}(\mathcal{G}, t) = \{R_l(i,j), \phi_d(i,j) \in \{(x,y), (x,z), (y,z), (x,t), (y,t), (z,t)\}, l \in \{1,2\}$, where $\mu = (x,y,z)$ is the mean of 3D Gaussians $\mathcal{G}$.

A multi-head Gaussian deformation decoder $\mathcal{D} = \{\varphi_x, \varphi_r, \varphi_s \}$ computes deformation components for position, rotation, and scaling, respectively, where each $\varphi$ is a separate MLP, leading to deformed features 
\begin{equation}
(\mathcal{X}', r', s') = (\mathcal{X}+\Delta\mathcal{X}, r+\Delta r, s + \Delta s)
\end{equation}
where $\Delta X = \varphi_x(\mathcal{X}), \Delta r = \varphi_r(r), \Delta s = \varphi_s(s)$. So the deformed 3D Gaussians are $\mathcal{G}' = \mathcal{G} + \Delta \mathcal{G} = \{\mathcal{X}', s', r', \sigma, \mathcal{C}\}$. Finally, a Gaussian conversion process based on timestamps $t_i$ maintains differential splatting efficiencies.

\subsection{CTRL-GS: Overview}
Key components of our method are depicted in Figure \ref{fig:method}. We build on the 4D-GS \cite{WYF+23} method for dynamic scenes.

At a high level, our cascaded decomposition is motivated by the real-world observation that dynamic scenes often exhibit hierarchical temporal structure: global geometry and lighting remain largely stable across the video, mid-level motion patterns (e.g. limb movement, deforming surfaces) evolve at an intermediate rate, and fine-grained changes (e.g. shading, occlusion boundaries) fluctuate on a per-frame basis. 

We first divide the frames of the dynamic scene into temporal windows. We experiment with three window construction methods: equal length, N-highest-flow, and dynamic thresholds. Next, we create reusable intermediate MLPs for position, rotation, and scaling, which are applied to the temporal windows and learn the deformation field from canonical 3D Gaussians for the frames within that window. Finally, our deformed features are computed as the sum of video-constant features, segment-specific means, and point-in-time deformations.

This "cascaded" approach allows us to capture changing scene information with greater flexibility and accuracy. We take advantage of information shared throughout the entire scene, while affording flexibility for appearing and disappearing elements and regions, large movements, and other temporal scene shifts. 

\subsection{Equal Temporal Window Construction}
Let timestamps $t_i$ within a scene fall within the range [0,1], and $N$ be the number of periods we select for temporal segmentation. Then the interval length $I = \frac{1}{N}$. We perform the following temporal feature transformation:
\begin{equation}
\mathcal{T} = \lfloor \frac{\mathcal{T}}{I+10^{-9}} \rfloor * I + q * I
\end{equation}
where $\mathcal{T}$ is an array containing the point timestamp of the same width as the other grid features and $q \in \mathbb{Q}+$ is the quantization coefficient, a performance hyperparameter that adjusts the interpolation between segment-level predictions and frame-specific residuals. We include $10^{-9}$ to ensure the floor function performs timestamp quantization as expected. Finally, we append this constructed temporal feature to the other existing grid features (representing $\mathcal{X}, r, s)$ to produce features with quantized temporal information $\mathcal{X}_\mathcal{T}, r_\mathcal{T}, s_\mathcal{T}$, as input to the segment-specific MLPs.

\subsection{Dynamic Window Construction: N-Highest-Flow }
Instead of simply dividing the frames into equal temporal windows, we can observe video characteristics and construct windows adaptively. Using a pretrained RAFT \cite{TD20} 'raft-things' model and one camera perspective, we calculate the optical flow $f_{ij}$ between each pair $i, j$ of consecutive frames in a dynamic scene. 

In each scene, we notice several 'peaks' of highest motion. These high-motion periods tend to divide a scene into different visual settings. Because models may have trouble learning over these high-motion periods, we use separate intermediate MLPs to learn before/after the moment of high motion. To yield $N$ temporal windows, we select the top $N-1$ frame-pairs $i,j$ with highest optical flow $f_{ij}$ as window divisions, so frame $i$ ends the previous window and $j$ begins the next.

\subsection{Dynamic Window Construction: Greedy Thresholds}

Extending this idea, we can try to construct more, shorter windows during periods of high motion where additional granularity can help learn significant geometry changes. We construct fewer, longer windows for periods of low motion where geometry is more temporally consistent. This approach allows us to incorporate more flexibility, leading to better reconstructions for high-motion scenes where current models tend to degrade most.

After selecting $N$ time periods for segmentation, we set threshold: 
\begin{equation}
T = \frac{\sum f_{ij}}{N}
\end{equation}
Iterating over the frames, we follow a greedy approach, keeping a sum of total flow so far. When we reach the threshold, we begin a new window and reset the flow sum.

\subsection{Cascaded Temporal Residue Deformation}
After constructing temporal windows from our dynamic scene with one of the three prior methods, we introduce three new MLPs $\varphi_{xT}, \varphi_{rT}, \varphi_{sT}$, which produce intermediate mean components for position, rotation, and scaling for each temporal segment individually: $\Delta_t \mathcal{X} = \varphi_{xT}(\mathcal X_T), \Delta_t r = \varphi_{rT}(r_T), \Delta_t s = \varphi_{sT}(s_T)$.

Now our deformed features are: 
\begin{equation}
(\mathcal{X}', r',s') = (\mathcal{X}+\Delta_t \mathcal{X} + \Delta \mathcal{X}, r + \Delta_t r + \Delta r, s + \Delta_t s + \Delta s)
\end{equation}

Finally, our deformed 4D Gaussians $\mathcal{G}'$ are defined by the updated features and the original opacity and spherical harmonic features: $\{ \mathcal{X}', s', r', \alpha, \mathcal{C} \}$. 

By using a "cascaded" approach, we model spatial-temporal features as the sum of video-constant features, segment-specific means, and point-in-time deformations. This approach offers greater reconstructive control over changing scenes.
\section{Experiments}

\begin{table}
\centering
\caption{Quantitative results on HyperNeRF \cite{PSH+21} vrig datasets with rendering resolution 960 x 540. First, second, and third best results are highlighted in \colorbox{yellow}{yellow},  \colorbox[HTML]{a3f3ff}{blue}, and \colorbox{pink}{pink}. CTRL-GS achieves the most improvements with dynamic threshold windows.}
\renewcommand\tabcolsep{3pt}
\begin{tabular}{l|c|c|c}
\toprule
\textbf{Model} & \textbf{PSNR (dB)} $\uparrow$ & \textbf{MS-SSIM} $\uparrow$ & \textbf{FPS} $\downarrow$ \\ \midrule
3D-GS \cite{KKL+23} &  19.7 & 0.680 & \cellcolor{yellow} 55            \\ 
Nerfies \cite{PSB+21} &  22.2 & 0.803 & $<$1            \\ 
HyperNeRF \cite{PSH+21} &  22.3 & 0.814 & $<$1           \\ 
FFDNeRF \cite{GSD+23} &  24.2 &  0.842 &  0.05            \\ 
TiNeuVox-B \cite{FYW+22} &  24.3  & 0.836 & 1             \\ 
V4D \cite{GXH+23} &  24.8 & 0.832 & 0.29             \\ 
4D-GS \cite{WYF+23} &  25.5 & \cellcolor{pink} 0.845 & \cellcolor[HTML]{a3f3ff} 34              \\ \hline
{\small Equal Windows} & \cellcolor{pink} 25.8 & 0.831 & \cellcolor{pink} 27 \\ 
{\small Dyn. N-highest} & \cellcolor[HTML]{a3f3ff} 25.9 & \cellcolor[HTML]{a3f3ff} 0.860 & 25 \\
{\small Dyn. Threshold} & \cellcolor{yellow} 26.0 & \cellcolor{yellow} 0.863 & 22 \\ 
\bottomrule
\end{tabular}
\label{tab:quant_results}
\end{table}

\begin{table*}
\centering
\caption{Quantitative results on D-NeRF \cite{PCP+21} synthetic datasets with rendering resolution 1352 x 1014. First, second, and third best results are highlighted in \colorbox{yellow}{yellow},  \colorbox[HTML]{a3f3ff}{blue}, and \colorbox{pink}{pink}. CTRL-GS with dynamic threshold-based temporal windows achieves the highest PSNR and SSIM metrics, but renders more slowly than 4D-GS.}
\begin{tabular}{l|c|c|c|c}
\toprule
\textbf{Model} & \textbf{PSNR (dB)}$\uparrow$ & \textbf{SSIM}$\uparrow$ & \textbf{LPIPS}$\downarrow$ & \textbf{FPS}$\uparrow$ \\ 
\midrule
TiNeuVox-B \cite{FYW+22}    & 32.67 & 0.97 & 0.04 & 1.5 \\ 
K-Planes \cite{FMR+23}       & 31.61 & 0.97 & 0.04 & 0.97 \\ 
HexPlane-Slim \cite{CJ23}   & 31.04 & 0.97 & 0.04 & 2.5 \\ 
3D-GS \cite{KKL+23}         & 23.19 & 0.93 & 0.12 & \cellcolor{yellow} 170 \\ 
FFDNeRF \cite{GSD+23}       & 32.68 & 0.97 & 0.04 & $<$1 \\ 
MSTH \cite{WCW+23}          & 31.34 & 0.98 & \cellcolor{pink}0.02 & 6 \\ 
V4D \cite{GXH+23}           & 33.72 & 0.98 & \cellcolor[HTML]{a3f3ff}0.02 & \cellcolor{pink}82 \\ 
4D-GS \cite{WYF+23}      & \cellcolor[HTML]{a3f3ff}34.05 & \cellcolor[HTML]{a3f3ff}0.98 & \cellcolor{yellow} 0.02 & \cellcolor[HTML]{a3f3ff} 82 \\ \hline
Ours: Equal Windows & 32.87 & 0.97 & 0.04 & 73 \\
Ours: Dynamic N-highest & \cellcolor{pink} 33.90 & \cellcolor{pink} 0.98 & 0.03 & 69\\ 
Ours: Dynamic Threshold & \cellcolor{yellow} 34.34 & \cellcolor{yellow} 0.98 & 0.03 & 63 \\
\bottomrule
\end{tabular}
\label{tab:quant_results_synthetic}
\end{table*}

\begin{table*}
\centering
\caption{Per-scene results on the HyperNeRF validation rig datasets \cite{PSH+21} across models. First, second, and third best results are highlighted in \colorbox{yellow}{yellow},  \colorbox[HTML]{a3f3ff}{blue}, and \colorbox{pink}{pink}, respectively. CTRL-GS achieves improved results across most datasets across multiple metrics.}
\renewcommand{\arraystretch}{1.2}
\begin{tabular}{l|cc|cc|cc|cc}
\toprule
\textbf{Method} & \multicolumn{2}{c|}{\textbf{3D Printer}} & \multicolumn{2}{c|}{\textbf{Chicken}} & \multicolumn{2}{c|}{\textbf{Broom}} & \multicolumn{2}{c}{\textbf{Banana}} \\ 
& PSNR & MS-SSIM & PSNR & MS-SSIM & PSNR & MS-SSIM & PSNR & MS-SSIM \\ 
\midrule
Nerfies \cite{PSB+21} & 20.6 & \cellcolor{pink} 0.83 & 26.7 & 0.94 & 19.2 & 0.56 & 22.4 & 0.87 \\ 
HyperNeRF \cite{PSH+21} & 20.0 & 0.59 & 26.9 & 0.94 & 19.3 & 0.59 & 23.3 & 0.90 \\ 
TiNeuVox-B \cite{FYW+22} & \cellcolor{yellow} 22.8 & \cellcolor{yellow} 0.84 & 28.3 & \cellcolor{yellow} 0.95 & 21.5 & 0.69 & 24.4 & 0.87 \\ 
FFDNeRF \cite{GSD+23} & \cellcolor[HTML]{a3f3ff} 22.8 & \cellcolor[HTML]{a3f3ff} 0.84 & 28.0 & \cellcolor{pink} 0.94 & 21.9 & 0.71 & 24.3 & 0.86 \\ 
3D-GS \cite{KKL+23} & 18.3 & 0.60 & 19.7 & 0.70 & 20.6 & 0.63 & 24.0 & 0.80 \\ 
4D-GS \cite{WYF+23} & 22.1 & 0.81 & 28.7 & 0.93 & 22.0 & 0.70 & 28.0 & \cellcolor{pink} 0.94 \\ \hline
Ours: Equal Windows & \cellcolor{pink} 22.2 & 0.74 & \cellcolor{pink} 29.1 & 0.93 & \cellcolor{pink} 22.3 & \cellcolor{pink} 0.72 & \cellcolor{pink}29.4 & 0.92 \\
Ours: Dynamic N-highest & 21.9 & 0.82 & \cellcolor[HTML]{a3f3ff} 29.5 & 0.93 & \cellcolor[HTML]{a3f3ff} 22.9 & \cellcolor[HTML]{a3f3ff} 0.75 & \cellcolor[HTML]{a3f3ff} 29.4 & \cellcolor[HTML]{a3f3ff} 0.94 \\
Ours: Dynamic Threshold & 21.9 & 0.82 & \cellcolor{yellow} 29.6 & \cellcolor[HTML]{a3f3ff} 0.94 & \cellcolor{yellow} 22.9 & \cellcolor{yellow} 0.75 & \cellcolor{yellow} 29.5 & \cellcolor{yellow} 0.94\\
\bottomrule
\end{tabular}
\label{tab:per_scene_hypernerf}
\end{table*}

\begin{table*}
\centering
\caption{Per-scene results on the synthetic datasets across models. First, second, and third best results are highlighted in \colorbox{yellow}{yellow},  \colorbox[HTML]{a3f3ff}{blue}, and \colorbox{pink}{pink}, respectively. CTRL-GS achieves improved results across most datasets across multiple metrics.}
\renewcommand\tabcolsep{4pt}
\begin{tabular}{l|ccc|ccc|ccc|ccc}
\toprule
\textbf{Method} & \multicolumn{3}{c|}{\textbf{Bouncing Balls}} & \multicolumn{3}{c|}{\textbf{Hellwarrior}} & \multicolumn{3}{c|}{\textbf{Hook}} & \multicolumn{3}{c}{\textbf{Jumpingjacks}} \\ 
 & PSNR & SSIM & LPIPS & PSNR & SSIM & LPIPS & PSNR & SSIM & LPIPS & PSNR & SSIM & LPIPS \\ 
\midrule
3D-GS      & 23.20 & 0.9591 & 0.0600 & 24.53 & 0.9336 & \cellcolor{pink}0.0580 & 21.71 & 0.8876 & 0.1034 & 23.20 & 0.9591 & 0.0600 \\ 
K-Planes   & 40.05 & \cellcolor{pink}0.9934 & \cellcolor{pink}0.0322 & 24.58 & 0.9520 & 0.0824 & 28.12 & \cellcolor{pink}0.9489 & \cellcolor{pink}0.0662 & 31.11 & 0.9708 & 0.0468 \\ 
HexPlane   & 39.86 & 0.9915 & 0.0323 & 25.00 & 0.9443 & 0.0732 & 28.63 & 0.9433 & 0.0636 & 33.49 & \cellcolor{pink}0.9772 & \cellcolor{pink}0.0398 \\ 
TiNeuVox   & \cellcolor{pink} 40.23 & 0.9926 & 0.0416 & \cellcolor{pink}27.10 & \cellcolor{pink}0.9638 & 0.0786 & \cellcolor{pink}28.63 & 0.9433 & 0.0636 & \cellcolor{pink}34.39 & 0.9771 & 0.0408 \\ 
4D-GS      & \cellcolor[HTML]{a3f3ff}40.62 & \cellcolor[HTML]{a3f3ff}0.9942 & \cellcolor[HTML]{a3f3ff}0.0155 & \cellcolor[HTML]{a3f3ff}28.71 & \cellcolor[HTML]{a3f3ff}0.9733 & \cellcolor[HTML]{a3f3ff}0.0369 & \cellcolor[HTML]{a3f3ff}32.73 & \cellcolor[HTML]{a3f3ff}0.9760 & \cellcolor[HTML]{a3f3ff}0.0272 & \cellcolor[HTML]{a3f3ff}35.42 & \cellcolor[HTML]{a3f3ff}0.9857 & \cellcolor{yellow}0.0128 \\  \hline
{\small Ours: Threshold} & \cellcolor{yellow}40.84 & \cellcolor{yellow}0.9946 & \cellcolor{yellow}0.0142 & \cellcolor{yellow}29.33 & \cellcolor{yellow}0.9742 & \cellcolor{yellow}0.0354 & \cellcolor{yellow}33.15 & \cellcolor{yellow}0.9765 & \cellcolor{yellow}0.0263 & \cellcolor{yellow}35.90 & \cellcolor{yellow}0.9862 & \cellcolor[HTML]{a3f3ff}0.0182 \\
\bottomrule

\toprule
\textbf{Method} & \multicolumn{3}{c|}{\textbf{Lego}} & \multicolumn{3}{c|}{\textbf{Mutant}} & \multicolumn{3}{c|}{\textbf{Standup}} & \multicolumn{3}{c}{\textbf{Trex}} \\ 
 & PSNR & SSIM & LPIPS & PSNR & SSIM & LPIPS & PSNR & SSIM & LPIPS & PSNR & SSIM & LPIPS \\ 
\midrule
3D-GS      & 23.06 & 0.9290 & 0.0642 & 20.64 & 0.9297 & 0.0828 & 21.91 & 0.9301 & 0.0785 & 21.93 & 0.9539 & 0.0487 \\ 
K-Planes   & \cellcolor{yellow}25.49 & \cellcolor{yellow}0.9483 & \cellcolor{yellow}0.0331 & 32.50 & 0.9713 & 0.0362 & 33.10 & 0.9793 & 0.0310 & 30.43 & 0.9737 & 0.0343 \\ 
HexPlane   & \cellcolor{pink}25.10 & \cellcolor[HTML]{a3f3ff}0.9388 & \cellcolor{pink}0.0437 & \cellcolor{pink}33.67 & \cellcolor{pink}0.9802 & \cellcolor{pink}0.0261 & 34.40 & \cellcolor{pink}0.9839 & \cellcolor{pink}0.0204 & \cellcolor{yellow}36.07 & \cellcolor{pink}0.9749 & \cellcolor{pink}0.0273 \\ 
TiNeuVox   & 24.65 & 0.9063 & 0.0648 & 30.87 & 0.9607 & 0.0474 & \cellcolor{pink}34.61 & 0.9797 & 0.0326 & 31.25 & 0.9666 & 0.0478 \\ 
4D-GS      & 25.03 & \cellcolor{pink}0.9376 & \cellcolor[HTML]{a3f3ff}0.0382 & \cellcolor[HTML]{a3f3ff}37.59 & \cellcolor[HTML]{a3f3ff}0.9880 & \cellcolor[HTML]{a3f3ff}0.0167 & \cellcolor[HTML]{a3f3ff}38.11 & \cellcolor[HTML]{a3f3ff}0.9898 & \cellcolor{yellow}0.0074 & \cellcolor[HTML]{a3f3ff}34.23 & \cellcolor[HTML]{a3f3ff}0.9850 & \cellcolor{yellow}0.0131 \\ \hline
{\small Ours: Threshold} & \cellcolor[HTML]{a3f3ff}25.23 & 0.9362 & 0.0611 & \cellcolor{yellow}38.16 & \cellcolor{yellow}0.9897 & \cellcolor{yellow}0.0143 & \cellcolor{yellow}38.15 & \cellcolor{yellow}0.9906 & \cellcolor[HTML]{a3f3ff}0.0126 & \cellcolor{pink}33.94 & \cellcolor{yellow}0.9858 & \cellcolor[HTML]{a3f3ff}0.0203 \\
\bottomrule
\end{tabular}
\label{tab:per_scene_synthetic}
\end{table*}

\subsection{Setting and Implementation Details}
We build on the implementation of 4D-GS \cite{WYF+23} using PyTorch. We use COLMAP \cite{SF16} to generate point clouds to initialize 3D Gaussians. To improve stability and convergence, we "warm up" optimization using smaller image resolution for 3000 initial iterations, which are then upsampled. We perform densification every 100 iterations until 10000 iterations, removing Gaussians with opacity $\alpha$ below threshold $\delta_\alpha$, which are essentially transparent. We observe 30,000 iterations for each scene and report best results for our models against baselines. All training and testing is performed on a single RTX A4000 GPU.

\subsection{Real-World Datasets}
We primarily assess the performance of our models on the 4 HyperNeRF \cite{PSH+21} vrig datasets, which contain real-world scenarios captured by two cameras. Each video, originally 30-60s long, has been sub-sampled to 15fps. The videos range approximately 300-1000 frames; the datasets are constructed by using every 4th frame for training, and the middle frame between each training pair for validation. 

\subsection{Synthetic Datasets}
We also evaluate our model on the 8 D-NeRF \cite{PCP+21} datasets, which contain monocular synthetic dynamic scenes without backgrounds. Each video contains 50-200 frames.

\subsection{Evaluation}

We primarily assess experimental results using metrics like peak-signal-to-noise ratio (PSNR), perceptual quality measure LPIPS \cite{ZIE+18}, structural similarity index (SSIM) \cite{WBS+04}, and its extension multiscale SSIM (MS-SSIM). We also compare our results with several other state-of-the-art methods as described previously, including Nerfies \cite{PSB+21}, HyperNeRF \cite{PSH+21}, FFDNeRF \cite{GSD+23}, TiNeuVox \cite{FYW+22}, V4D \cite{GXH+23} , K-Planes \cite{FMR+23}, HexPlane \cite{CJ23}, MSTH \cite{WCW+23}, 3D Gaussian Splatting \cite{KKL+23}, and 4D Gaussian Splatting \cite{WYF+23}. Results for comparison methods are from their papers, code, or otherwise provided by their authors. Rendering speed is borrowed from estimates by \cite{WYF+23} on official implementations.

\subsection{Quantitative Results}
Results on the real-world datasets are summarized in Tab. \ref{tab:quant_results}, and enumerated per scene in Tab. \ref{tab:per_scene_hypernerf}. Results on the synthetic datasets are summarized in Tab. \ref{tab:quant_results_synthetic}, and enumerated per scene in Tab. \ref{tab:per_scene_synthetic}. CTRL-GS shows improved accuracy across scenes, with the greatest improvements occurring in scenes with high motion, occluded areas, or fine detail that is challenging for current models to reconstruct. For example, on the \textsc{Broom} sequence–characterized by all of the aforementioned challenges, CTRL-GS achieves 22.9 PSNR, outperforming all cited methods that presented scene-level results. These include (in addition to the results collated in the tables) DynMF \cite{KLD24} (22.1), Gaussian-Flow \cite{LDZ+24} (22.8), and SP-GS \cite{WLZ24} (22.8).

Like other Gaussian Splatting models, CTRL-GS enjoys much higher rendering speed than NeRF-based methods. We observed some slowdown from 4D-GS, which is primarily due to repeated segment identification during rendering, which was introduced to support our dynamic segment-based querying. Specifically, in the forward-dynamic method, we perform per-frame checks to determine segment membership. The underlying issue is not the segment strategy itself, but rather the inefficiency of dynamically querying segment info on-the-fly. This can be fixed by precomputing segment indices during training and caching them for efficient lookup during rendering. We plan to implement this improvement for the official release.

\subsection{Qualitative Results}
Qualitative results are presented in Fig. \ref{fig:hypernerf_images} for real-world datasets and \ref{fig:dnerf_images} for synthetic datasets. Our method can more accurately reconstruct structure, handle edges more smoothly, recover more accurate color and lighting, and preserve fine details. For example, in HyperNeRF-\textsc{Banana}, CTRL-GS correctly places fingers where 4D-GS constructs a thumb not present in the ground truth. In \textsc{Chicken}, CTRL-GS correctly constructs a smooth lid and clear fingernail contours, where 4D-GS produces incorrect lid holes and fingertip splintering. In D-NeRF-\textsc{Mutant}, CTRL-GS captures the fine details of dark cracks, blue veins, and surface textures, also preserving original luminosity. Note that 4D-GS shows significant blurring both within and on the edges of the structures. In \textsc{Standup}, CTRL-GS replicates the crisp and precise helmet contours, more accurate stripe texture, and beard shape, where 4D-GS performs incorrect texture and shape blurring throughout. In \textsc{Hellwarrior}, our method correctly constructs the spheres on the armor, upper breastplate segmentation, highlight luminosity and shadow depth, which are missed by 4D-GS. These improvements demonstrate an improved ability to capture realism, especially in scenes with fine details and textures. Additional comparison frames can be found in the appendix. 

\subsection{Ablation Studies}
We compare our three temporal window construction methods and find that the greedy threshold-based dynamic method tends to perform best across settings. However, even simpler methods like equal temporal windows and N-highest windows offer improvements from baseline models.

\subsection{Limitations}
While CTRL-GS does not uniformly surpass all existing methods across every dataset, it delivers notable and consistent improvements on several key benchmarks and scenes (for example, HyperNeRF's \textsc{Broom}). In general, CTRL-GS offers the most meaningful advances on challenging dynamic scenes where range of motion is high and/or complex occlusions are present, cases where existing approaches degrade most (often producing significantly inaccurate reconstructions with holes, visual sharding, temporal incontinuity, and ghosting). The benchmark datasets offer a range of scenes with different characteristics. In scenes where motion is limited and temporal variation is low, existing models already perform reasonably well, and improvements offered by CTRL-GS are not as significant. 

\section{Conclusion}

This paper proposes Cascaded Temporal Residue Learning for 4D-GS (CTRL-GS) to improve visual fidelity in reconstruction tasks. We evaluate our models across a wide set of real-world and synthetic data, with different kinds of motion, lighting, and occlusions, and find that CTRL-GS produces the greatest improvements on challenging settings where there is high motion throughout the temporal sequence, where current models struggle most, often producing significantly inaccurate reconstructions with holes, visual sharding, temporal incontinuity, and ghosting. Future work may explore 1) further reducing aberrations, 2) static-dynamic decomposition, 3) adaptive opacity controls for objects entering and leaving the scene, and 4) improving rendering speed.

\section*{Acknowledgments}
This work is supported in part by NIH grant R01HD104969.

{
    \small
    \bibliographystyle{ieeenat_fullname}

}

\clearpage
\setcounter{page}{1}
\maketitlesupplementary

\section*{Optimization Details}
We mostly follow the same settings as used in 4D-GS \cite{WYF+23}. We observe 30,000 iterations for each scene and report best results for our models against baselines.

To test each temporal window construction method, we perform a hyperparameter sweep over number of windows $N \in \{2, 3, 4, 5, 6, 7, 8, 9\}$ for all methods, and quantization coefficient $q \in \{0.0, 0.1, 0.2, 0.3, 0.5, 0.7, 0.9\}$ for the equal-length method, reporting best results. This set of values for $q$ was selected to test a range of interpolation weights between segment and residual predictions. We did not choose to test higher numbers of $N$, because we tended to note performance degradation at the upper range of this set (which makes sense, because for short videos, too many temporal segments can lead to the consequences of temporal discontinuity outweighing gains from temporal fitting). 

\begin{figure}
\centering
\includegraphics[width=\columnwidth]{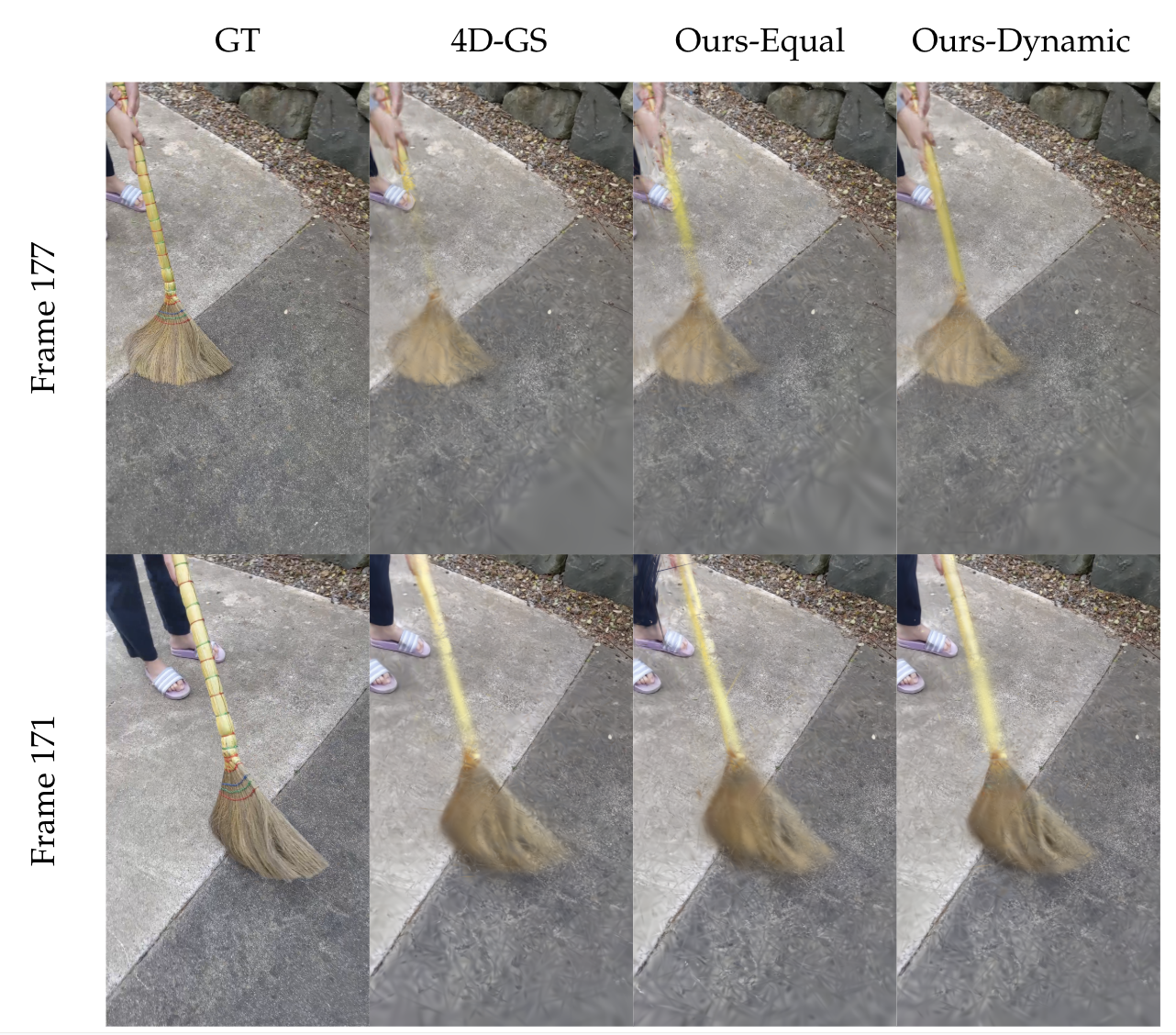}  
\caption{Additional frame comparisons for the HyperNeRF \cite{PSH+21} \textsc{Broom} dataset across ground truth (GT), baseline 4D-GS, CTRL-GS with equal temporal windows, and CTRL-GS with dynamic threshold-based windows. In frame 177, our model shows a full handle reconstruction. In frame 171, our model shows more accurate red/green coloring, depth, and fiber detail.}
\label{fig:broom}
\end{figure}

\begin{figure}[H]
\centering
\includegraphics[width=\columnwidth]{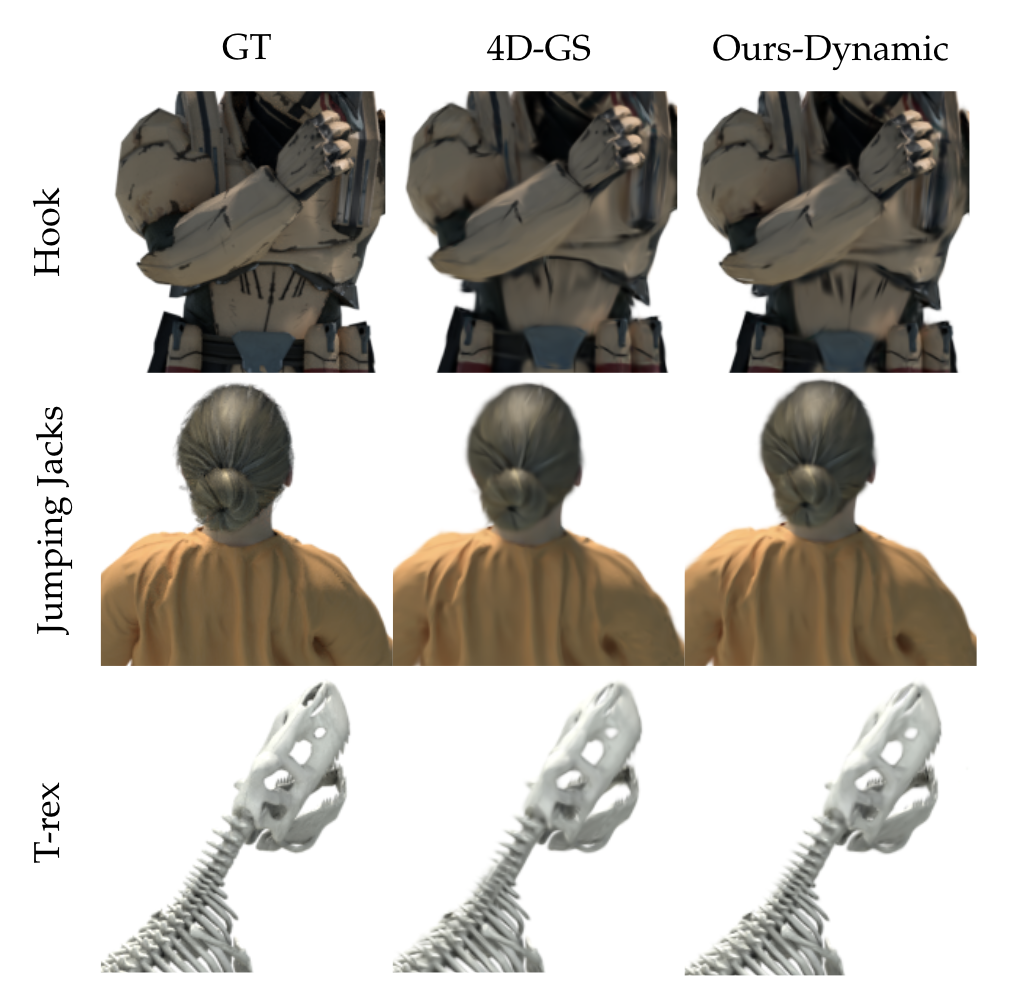}  
\caption{Frame comparisons for the D-NeRF \cite{PCP+21} \textsc{Hook}, \textsc{Jumping Jacks}, and \textsc{T-Rex} datasets across ground truth (GT), baseline 4D-GS, and CTRL-GS with dynamic threshold-based windows. CTRL-GS more accurately reconstructs fine details, like armor lines, hair texture, and T-rex teeth.}
\label{fig:dnerf2}
\end{figure}


\end{document}